# A Bidirectional Conversion Network for Cross-Spectral Face Recognition

Zhicheng Cao, Jiaxuan Zhang and Liaojun Pang

Molecular and Neuroimaging Engineering Research Center of Ministry of Education, School of Life Science and Technology, Xidian University, Xi'an, China 710071.

## Abstract

Face recognition in the infrared (IR) band has become an important supplement to visible light face recognition due to its advantages of independent background light, strong penetration, ability of imaging under harsh environments such as nighttime, rain and fog. However, cross-spectral face recognition (i.e., VIS to IR) is very challenging due to the dramatic difference between the visible light and IR imageries as well as the lack of paired training data. This paper proposes a framework of bidirectional cross-spectral conversion (BCSC-GAN) between the heterogeneous face images, and designs an adaptive weighted fusion mechanism based on information fusion theory. The network reduces the cross-spectral recognition problem into an intra-spectral problem, and improves performance by fusing bidirectional information. Specifically, a face identity retaining module (IRM) is introduced with the ability to preserve identity features, and a new composite loss function is designed to overcome the modal differences caused by different spectral characteristics. Two datasets of TINDERS and CASIA were tested, where performance metrics of FID, recognition rate, equal error rate and normalized distance were compared. Results show that our proposed network is superior than other state-of-the-art methods. Additionally, the proposed rule of Self Adaptive Weighted Fusion (SAWF) is better than the recognition results of the unfused case and other traditional fusion rules that are commonly used, which further justifies the effectiveness and superiority of the proposed bidirectional conversion approach.

# 1. Introduction

Face recognition has become one of the most widely used biometric recognition technologies because of its universality and non-invasiveness, and is therefore favored by industry and academia with continuous progress being made [1]. With the advent and development of deep learning technology, face recognition has achieved very high accuracy under good illumination conditions [2, 3].

However, there still exist unsolved challenges with the problem of face recognition. For example, most of current face recognition technologies use the visible light (Vis) band, and its performance will be significantly reduced under unstable ambient illumination or complex illumination conditions such as nighttime. The infrared (IR) band of the electromagnetic spectra has the advantages of low background illumination requirement, strong penetration and ability of imaging in rainy and foggy weather, which can serve as supplement to the visible light band. Therefore, the use of IR face recognition technology is particularly urgent in practice. In recent years, Vis-NIR heterogeneous face recognition has been paid more and more attention by researchers and enterprises due to its application value in security monitoring [4], e-passport [5] and other practical scenarios.

VIS-IR heterogeneous face recognition is a far more challenging task than the intra-spectral recognition problem under the visible light band alone, due to the dramatic difference between the visible light and IR face images, as well as the lack of paired training data. The problem of cross-spectral face recognition is more complex because it needs to solve how to extract common face features under different spectra. Early research works mostly focused on the near-infrared band, and mostly used the near-infrared single band for face recognition within the spectrum, that is, the training set and test set for the same band under the data collection. This kind of work is essentially a simple extension of the traditional visible face recognition. For example, Chen et al. [8] and Li et al. [9] used the near-infrared band of infrared to explore face recognition within the same spectrum. Following up, scholars used multiple infrared wave bands such as the near infrared and medium wave infrared. For instance, Kong [10] and Singh [11] have tried the new research perspective of multiband fusion. But the corresponding recognition problem is still essentially an intra-spectral recognition problem.

It was scholars such as Jain [12] and Dong [13] who really began to study cross-spectral face recognition. Up to now, cross-spectral face recognition methods can be divided into two categories: the traditional feature extraction methods and the deep learning-based methods. The traditional feature extraction method is based on manual design and can be further divided into global method and local feature method. The former uses subspace projection to analyze the global illumination measurement information of the face and find the closest pattern, such as Principal Component Analysis (PCA) and Linear Discriminant Analysis (LDA). The latter uses local operators and descriptors to extract local attributes, such as Local Binary Pattern (LBP) and Histogram of Oriented Gradient (HOG). Xu et al. [14] used the sparse representation technology to construct a joint cross-spectral dictionary to find new ideas for cross-spectral face recognition and achieved good results. Gong et al. [15] used the similar probability LDA method to encode and then train the generic differentiation model on NIR and visible images to extract the heterogeneous common features. Local feature method uses the idea of local operators. For example, Jain et al. [12] used popular feature extraction operators such as LBP and HOG to extract face features of near infrared and visible light respectively, and conducted cross-spectral comparison between the two. Cao et al. [16] integrated LBP and WLD features and introduced smooth composite multi-lobe operators, and designed face comparison experiments between short-wave infrared and visible light at varying distances. However, in general, traditional feature extraction methods are gradually replaced by deep learning methods due to issues of complex design and poor robustness.

More recently, convolutional neural network and other deep learning technologies have increasingly shown their performance advantages over traditional methods [17]. So far, cross-spectral face recognition methods based on deep learning can be divided into the following two categories: the invariant domain feature methods and the image synthesis and conversion methods. The first type of methods is based on potential subspace learning method, which projects extracted features from two different fields onto a common potential subspace and carries out feature matching. For example, Riggan et al. [18] proposed a coupling automatic correlation neural network. By forcing the potential features of the two neural networks to be as similar as possible, face images of different forms are projected into the potential space, while retaining the information from the input, learning cross-mode transformation, and comparing feature vectors for face recognition. Le et al. [19], based on the domain-invariant feature learning method, introduce domain-based Angle edge loss and maximum Angle loss to reduce domain information, learn distinguishing features, maintain differences between classes, and strengthen the close relationship between labels in the dataset.

With the rapid development of deep learning tools, it provides an opportunity for us to study cross-spectral face recognition. Among them, the most eye-catching work is image synthesis and transformation, such as Variational Auto Encoder [6], Generative Adversarial Network [7]. Researchers have applied them to the cross-spectral face recognition problem. For example, Riggan et al. [20] project the data of one mode into the space of another mode, so as to measure the similarity between heterogeneous data in different domains. Song et al. [21] integrated cross-spectral face generation and discriminant feature learning into the end-to-end adversarial network, adopted a two-path model to learn global structure and local texture in pixel space, and adopted adversarial loss and high-order variance loss to measure the overall and local differences of two heterogeneous distributions in feature space.

However, most of these deep generation technologies are designed for image style transfer, and not for face images. Therefore, the effect is not optimal for cross-spectral face recognition and the technology does not ensure the preservation of identity features -- and identity preservation is crucial in face recognition. What's more, most of these networks for face image generation are limited to one-direction face generation and fail to consider the case of bidirectional face conversion. Therefore, direct utilization of these methods cannot well meet the requirement of bidirectional cross-spectral face conversion in our cross-spectral face recognition problem.

In view of this, we in this paper proposes a bidirectional conversion framework for cross-spectral face recognition, Bidirectional Cross-Spectral Conversion GAN (BCSC-GAN). A face feature extraction module is introduced and a new composite loss function is designed to overcome the modal differences caused by different spectra to obtain better results. The proposed network conducts face image conversion from visible light to IR and vice versa while extracting and preserving the identity information of heterogenous face images, such that the network reduces the cross-spectral recognition problem into an intra-spectral problem. After bidirectional face image conversion, the proposed method improves the cross-spectral face recognition performance by fusing bidirectional matching scores. During training, the model of FaceNet [2] is pre-trained on the WebFace face databset and then is transfer-learned uses the CASIA and TINDERS face datasets. During designing of the score fusion rule, instead of manual or fixed weights as in other common fusion rules, we adopt the idea of adaptive weighted fusion mechanism which introduces face recognition confidence index. Contributions of this paper are listed as follows:

(1) A new cross-spectral face recognition method is proposed based on image transformation. Our proposed method features bidirectional cross-spectral face image conversion. The bidirectional conversion approach is fundamentally better than other SOTA methods of single directional conversion because bidirectional matching scores can be fused to improve the final recognition accuracy.

(2) The proposed network (BCSC-GAN) not only achieves high quality face conversion effect but also ensures the preservation of face identity information during cross-spectral face image conversion, which is crucial to improvement of the recognition performance. Compared with other methods, it has higher recognition performance and stronger robustness.

(3) A new composite loss function is designed to effectively overcome the modal differences caused by different spectral characteristics, so as to obtain better generation results. The composite loss function comprehensively considers adversarial loss, cycle-consistency loss, synthesized loss and identity retaining loss.

(4) A self-adaptive weighted fusion (SAWF) mechanism is designed in which the GAR and d-prime values of single spectral matching scores are used to automatically update the weights. Recognition accuracy using the SAWF is better than that of the non-fusion and that of the traditional fusion algorithms.

The rest of the paper is organized as follows: In Section 2, we first introduce the overall structure of our proposed network, and then explain the proposed composite loss function as well as the information fusion strategy. Section 3 introduces the datasets for testing, the performance evaluation metrics and the experimental setups, and provides experimental results and analysis. The last section summarizes the findings and contributions of this research work.

## 2. Proposed Method

This paper proposes a network framework of bidirectional face image conversion for cross-spectral face recognition. Such network successfully reduces the problem of cross-spectral face recognition into a single spectrum matching problem. A face feature extraction module with high differentiation ability is introduced to extract common and heterogeneous face features, and a new composite loss function is designed to overcome the modal differences caused by different spectral characteristics. In order to further improve the accuracy of cross-spectral face recognition, a self-adaptive weighted fusion mechanism is proposed to fuse the matching scores under each light band.

**2.1 Network Structure**

In this paper, a bidirectional conversion Generative Adversarial Network model BCSC-GAN is proposed and constructed. The network includes generator module *G*, generator module *F*, visible discriminator module $D_{VIS}$, infrared discriminator module $D_{IR}$, and identity retaining module IRM. Figure 1 shows the overall structure of the BCSC-GAN network.

The generator modules of *G* and *F* are both in the form of the encoder-decoder network, and consist of down-sampling and up-sampling layers of the same number. The network comprises a convolutional layer with a convolutional core of 7*7, 2 convolutional layers with 3*3 convolutional cores, 9 residual blocks, 2 convolutional kernels with 3*3 convolutional cores, and 1 convolutional layer with a convolutional kernel of 7*7. The output of each convolutional layer is regularized by instances, where the first half is the image encoding process and the second half is the image decoding process. In order to reduce the artifacts of the input image and the generation images during the convolution process, layers 1 and 15 are filled with reflection padding of 3, and padding is filled with 1 in other layers. The specific network parameters are shown in Table 1.

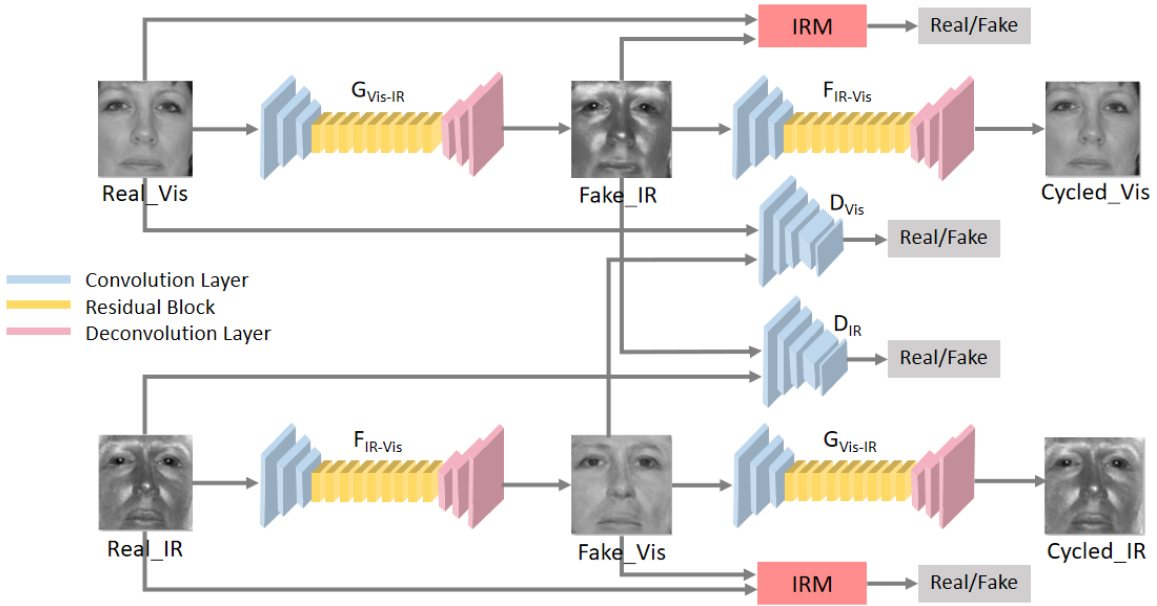

Figure 1. The structure of the proposed BCSC-GAN network for bidirectional cross-spectral face conversion.

Table 1. Specific network parameters for the generator of BCSC-GAN.

| Module | Layer | Type | Step Size | Convolution kernel number | Input channel number | Output channel number |
|---|---|---|---|---|---|---|
| Encoder | 1 | 7×7 conv - Instance normalization -ReLU | 1 | 64 | 1 | 64 |
| | 2 | 3×3 conv - Instance normalization -ReLU | 2 | 128 | 64 | 128 |
| | 3 | 3×3 conv - Instance normalization -ReLU | 2 | 256 | 128 | 256 |
| Image conversion core | 4 | Two 3×3conv Residual block- Instance normalization -ReLU | 1 | 256 | 256 | 256 |
| | 5 | Two 3×3conv Residual block- Instance normalization-ReLU | 1 | 256 | 256 | 256 |
| | 6 | Two 3×3conv Residual block - Instance normalization-ReLU | 1 | 256 | 256 | 256 |
| | 7 | Two 3×3conv Residual block - Instance normalization-ReLU | 1 | 256 | 256 | 256 |
| | 8 | Two 3×3 conv Residual block- Instance normalization-ReLU | 1 | 256 | 256 | 256 |
| | 9 | Two 3×3conv Residual block - Instance normalization-ReLU | 1 | 256 | 256 | 256 |
| | 10 | Two 3×3conv Residual block - Instance normalization-ReLU | 1 | 256 | 256 | 256 |
| | 11 | Two 3×3conv Residual block - Instance normalization-ReLU | 1 | 256 | 256 | 256 |
| | 12 | Two 3×3conv Residual block - Instance normalization-ReLU | 1 | 256 | 256 | 256 |
| Decoder | 13 | 3×3 deconv-Instance normalization-ReLU | 1/2 | 128 | 256 | 128 |
| | 14 | 3×3 deconv-Instance normalization-ReLU | 1/2 | 64 | 128 | 64 |
| | 15 | 7×7conv-Instance normalization-ReLU | 1 | 1 | 64 | 1 |

The discriminator module takes the Markov Patch GAN structure [37]. Compared with the input of the whole image for discrimination, the patch of 70×70 pixels can effectively capture local high-frequency features, such as texture features. The discriminator consists of four convolution layers with 4×4 step lengths of 2, and the final result is output by a convolution layer. The network uses Leaky ReLU with slope of 0.2 as the activation function. Input images to the discriminator are in the size of 256×256 pixels. The output is the discrimination probability of true or false for each given input image, which is between 0 and 1. Specific network parameters are shown in Table 2.

Table 2. The specific network parameters for the discriminator of BCSC-GAN.

| Layer | Type | Step size | Convolution kernels number | Input channel number | Output channel number |
|---|---|---|---|---|---|
| C1 | 4×4conv-LReLU | 2 | 64 | 1 | 64 |
| C2 | 4×4conv-Instance normalization-LReLU | 2 | 128 | 64 | 128 |
| C3 | 4×4conv-Instance normalization-LReLU | 2 | 256 | 128 | 256 |
| C4 | 4×4conv-Instance normalization-LReLU | 1 | 512 | 256 | 512 |
| C5 | 4×4conv | 1 | 1 | 512 | 1 |

The concept of identity preservation was originally proposed to solve the problem of position invariance. Such as, Liu, etc. [38] designed a network of semantic discriminator for cross-domain pedestrian recognition, and kept the posture information of the source domain image during the image conversion process. Using the semantic constraints of identity loss function, the identity of the cross-domain pedestrians is kept in the identification problem. Wang et al. [39] designed and introduced a pre-trained model to extract high-dimensional features from generated images and real images respectively, and proposed a perceptual loss function to measure the perceptual details of the extracted middle layer, so as to maintain identity information. Inspired by the above literature, this paper proposes an Identity Retaining Module (IRM) specifically for face recognition tasks, at the bidirectional cross-spectral face conversion stage.

The IRM module is designed in a succession of three steps: the face feature extraction step, the feature normalization step, and the feature space mapping step, as shown in Figure 2. After feature extraction and normalization of two input face images using IRM network structure, the L2 distance is calculated in the Euclidean space. The L2 distance is then used as the identity retention loss (i.e., one of the composite loss terms) to update network parameters and optimize the network during back propagation. The face feature extraction model is trained and saved on the basis of the pre-trained model of FaceNet and then by transfer learning on the CASIA dataset, such that it can efficiently extract face features and compete with the generator under the constraint of the identity retention loss.

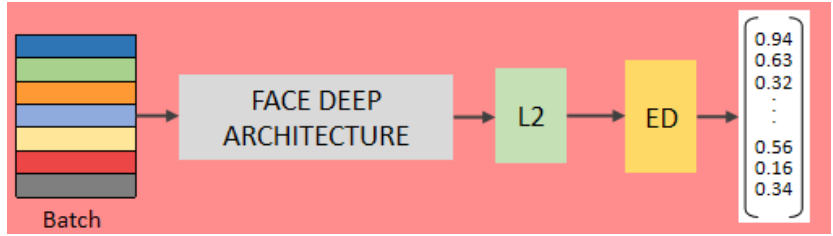

Figure 2. The structure of the Identity Retaining Module.

## 2.2 Composite loss Function

After the design of bidirectional image conversion network structure, the following critical problem is to design an effective loss function. At present, the importance and effect of loss function on model performance is often ignored in image generation/synthesis techniques. Most common loss functions are adversarial loss and $L_1$ loss, but there is no loss function especially proposed for the face recognition task. In order to ensure the generated face is in high quality and the final face recognition is well-performing in the meantime, it is necessary and crucial to preserve the facial details and spectral information useful for face recognition task as accurately as possible during image generation. Therefore, a composite loss function consisting of adversarial loss, cycle-consistency loss, synthesized loss and identity retaining loss is introduced by us to train the generators and discriminators, as shown in Figure 3.

The mathematical expression of the proposed composite loss function is given as follows:

$$Loss = L_{GAN} + \lambda_{cyc}L_{cyc} + \lambda_{syn}L_{syn} + \lambda_{IDR}L_{IDR} \quad (1)$$

where $L_{GAN}, L_{cyc}, L_{syn}$ and $L_{IDR}$ represent the adversarial loss, the cycle-consistency loss, the synthesized loss and the identity retaining loss, respectively. The coefficient $\lambda$ is used to adjust the proportion of each loss term in the total loss. In the following experiments, the values are empirically set to be $\lambda_{cyc}=10$, $\lambda_{syn}=30$, $\lambda_{IDR}=10$, respectively.

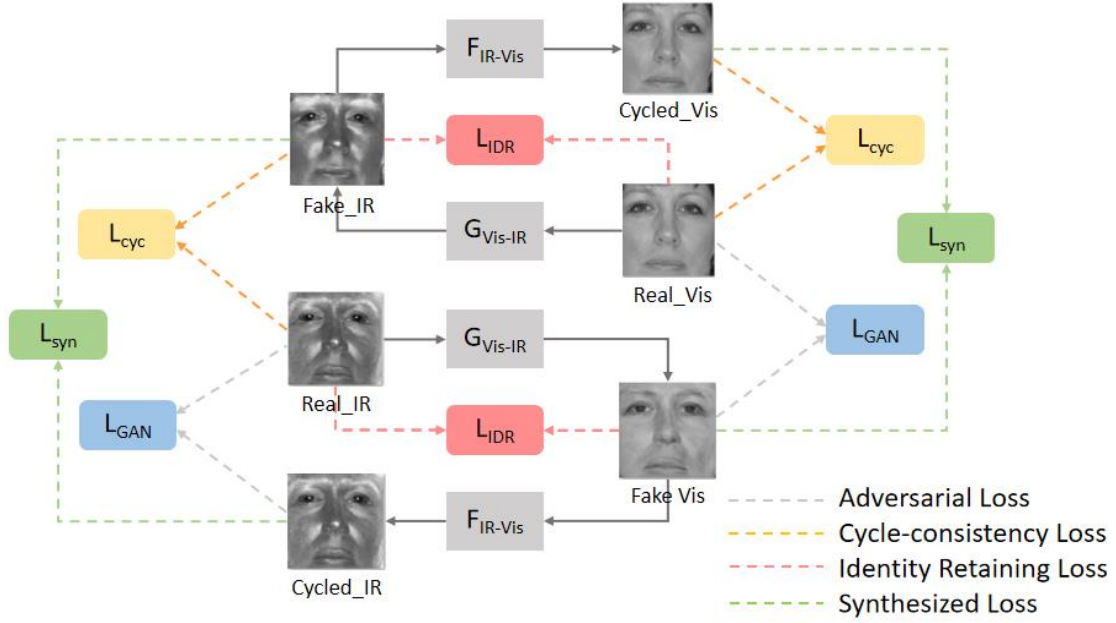

Figure 3. Design of the composite loss function.

The first loss term, $L_{GAN}$, serves as the foundation of the composite loss function of the whole network. By calculating the probability estimation of the real input face sample and the generated face sample by the discriminator, it helps construct high frequency details in a game-competing manner to ensure the accuracy of the high frequency components, so as to improve the generation efficiency. The expression of $L_{GAN}$ is as given by:

$$L_{GAN}(G,F,D_{VIS},D_{IR}) = L_{GAN}(G,D_{IR},V,I) + L_{GAN}(F,D_{VIS},I,V)$$
$$= E_{i \sim Pdata(I)}[\log D_{IR}(i)] + E_{v \sim Pdata(V)}[\log(1 - D_{IR}(G(v)))] +$$
$$E_{v \sim Pdata(V)}[\log D_{VIS}(v)] + E_{i \sim Pdata(I)}[\log(1 - D_{VIS}(F(i)))] \qquad (2)$$

where $V$ and $I$ represent the input image set for the visible light and the IR, respectively, in the bidirectional cross-spectral image conversion network. $v$ and $i$ represent a Vis and IR face sample, respectively. $L_{GAN}(G,D_{IR},V,I)$ is calculated as the cross-entropy loss between the discriminant outputs of $D_{IR}$ on $i$ and the generated result of $v$ through $G$. Similarly, $L_{GAN}(F,D_{VIS},I,V)$ is calculated as the cross-entropy loss between the discriminant outputs of $D_{VIS}$ on $v$ and the generated result of $i$ through $F$. During training, generators $G$ and $F$ try the best to minimize the target while, on the contrary, discriminators $D_{IR}$ and $D_{VIS}$ try to maximize the target utmost. The generators and discriminators are optimized alternately, such that the image generated by the generator is as close as possible to the real image.

Since two generators are involved in bidirectional network (one for Vis and another for IR), the cycle-consistency loss, $L_{cyc}$, is also designed to include two terms which ensures approximation of the cyclically generated image to the original input image as well as the accuracy of low-frequency components. In this way, a high-precision generator is trained which correctly learns the mapping relationship. Its mathematical expression is as follows:

$$L_{cyc}(G,F) = E_{v \sim Pdata(V)}[\|F(G(v)) - v\|_1] + E_{i \sim Pdata(I)}[\|G(F(i)) - i\|_1] \quad (3)$$

where $\|\bullet\|_1$ represents the $L_1$ norm.

$L_{syn}$ is the synthesis loss term, which ensures that the generated face image is as close to the cyclically synthesized face image as possible by calculating the $L_1$ distance between the cyclically synthesized image and the generated image in the same light band. It is calculated as:

$$L_{syn}(G,F) = E_{v \sim P(V), i \sim P(I)}[\|F(i) - F(G(v))\|_1] + E_{v \sim P(V), i \sim P(I)}[\|G(v) - G(F(i))\|_1] \quad (4)$$

where $L_{syn}$ is composed of two terms: the $L_1$-norm loss between the generated image of $i$ through $F$ and the generated image of $v$ through $G$ and then through $F$, and the $L_1$-norm loss between the generated image of $v$ through $G$ and the generated image of $i$ through $F$ and then through $G$.

$L_{IDR}$ is the identity retaining loss proposed in this paper, which is meant to overcome the problem of many image generation algorithms -- they only consider the quality of image generation but ignore the recognition purpose after image generation. Therefore, these algorithms usually perform well in terms of image generation quality but is unsatisfactory in terms of recognition rate. Thus, we in this paper introduce identity retaining loss especially for the context of cross-spectral face recognition, which not only ensures the integrity and consistency of identity information in the process of bidirectional face conversion, but also ensures high recognition performance. It is mathematically defined as:

$$L_{IDR} = \sqrt{\sum_{m=1}^{128}[emb_{VIS}^{(m)} - emb_{IR}^{(m)}]^2} \quad (5)$$

where the original face image and the generated cross-spectral face image were input to the face feature extraction network at the same time, and the corresponding feature vectors $\{emb_{VIS}^{(m)}\}$ and $\{emb_{IR}^{(m)}\}$ with 128-dimensions were extracted. Then, the Euclidean distance was calculated between these two features as their loss values were propagated back to the network, which further constrained and optimized the performance of generator and discriminator.

**2.3 Score-Level Information Fusion**

Since our proposed approach of bidirectional face conversion generates both IR

and Vis faces against their opponents, two-way matching scores are available. Therefore, we can further improve the cross-spectral recognition rate by fusing the two-way matching scores. According to the process and mode of information fusion, information fusion can be divided into sensor level [40-42], feature level [43-45], score level and decision level [46]. Among the four levels of fusion, score fusion is realized by fusion of the matching scores of different sources. Because it contains neither too much redundant information nor insufficient information required for identification of a person, so in many biometric systems score level fusion is superior to the other levels of fusion and is the mostly applied and studied fusion method at present.

Score level fusion methods can be further divided into three categories: fusion algorithms based on statistical normalization rules, density-based fusion algorithms and classifier-based fusion algorithms [47] . Common score fusion rules based on statistical normalization matching include simple rules such as calculation of the sum, product, maximum, minimum and median [48], and other more complex rules such as geometric average, arithmetic average and weighted average. Indovina et al. [49] and Wang et al. [50] proposed the weighted sum rule in which the weights were calculated according to the equal error rate index of each simple classifier. Kabir et al. [51] proposed a weighting rule based on the confidence of matching score by using the overlapping neighborhood between real matching score and deception score.

In this paper, we propose a score level fusion strategy based on statistical normalization and refer it as Self-Adaptive Weighted Fusion (SAWF), using the GAR and d-prime values of the bidirectional matching process. The SAWF fusion strategy is realized as follows: The GAR and d-prime values are first calculated from the two matching matrices (i.e., Vis-to-IR matching and IR-to-Vis matching) and are then used as weight coefficients of the matching scores to automatically update the weights. The updated weights are finally used to add the matching scores to fuse the information of visible and infrared face matching matrices. The mathematical process of SAWF is given as follows:

If $d - prime_V > d - prime_I$, then set:

$$\begin{cases} w_1 = G_V/(G_V + G_I) \\ w_2 = 1 - w_1 \end{cases} \qquad (6)$$

Else if $d - prime_V < d - prime_I$, then set:

$$\begin{cases} w_2 = G_I/(G_I + G_V) \\ w_1 = 1 - w_2 \end{cases} \qquad (7)$$

Else if $d - prime_V = d - prime_I$, then set:

$$w_1 = w_2 = \frac{1}{2} \tag{8}$$

The final fused matching score is set to be:

$$MatchScore = w_1 \times MatchScore_V + w_2 \times MatchScore_I \tag{9}$$

where $G_V$ and $d-prime_V$ denote the GAR value and d-prime value for the visible light matching score, respectively; $G_I$ and $d-prime_I$ represent the GAR value and d-prime value for the IR face matching score, respectively; $MatchScore_V$ and $MatchScore_I$ stand for the visible face matching score and IR matching score, respectively; $w_1$ and $w_2$ represent the weights calculated by the fusion formula during the process of score fusion.

Utilizing the GAR values of the two matching matrices, the above fusion strategy can be used to calculate the fusion weights adaptively, and the fusion weight of the matching matrix with smaller d-prime values can be guaranteed to be greater than 1/2, so as to improve the fusion and matching performance.

## 3. Experimental Results

Our proposed method of BCSC-GAN is compared with other state-of-the-art methods in the literature, such as image generation method based on shared potential subspace (CpAE)[18], Pix2pix[26], CycleGAN[29] and CSGAN[24]. The Genuine Acceptance Rate (GAR), Equal Error Rate (EER), d-prime and Fréchet Inception distance (FID) of the bidirectionally generated face image and the original face image of corresponding light bands were calculated. Then, score level fusion experiments were carried out where our SAWF rule was compared with other traditional fusion rules such as the maximum rule, geometric average rule and arithmetic average rule to obtain the final result of cross-spectral face recognition.

### 3.1 Datasets and Experimental Setups

In order to justify our bidirectional conversion approach and our fusion strategy, we carry out cross-spectral face recognition experiments and compare the performance of our approach against that of other SOTA methods. Throughout the experiments, we use two cross-spectral face datasets, the CASIA NIR-VIS 2.0 dataset and the Tactical Imager for Night/Day Extended-range Surveillance dataset (TINDERS).

The CASIA dataset was collected and sorted out by the Institute of Automation, Chinese Academy of Sciences, and was divided into four subsets S1~S4, each of which was collected at different times and under different light conditions, containing 17,580 natural expression frontal face images of 725 Asians. The TINDERS dataset consists of 1255 frontal face images of 48 individuals, which were collected in visible light (VIS), near infrared (NIR) and short-wave infrared (SWIR) scenarios. Since the size of the input image of the cross-spectral face bidirectional conversion network designed in this paper is 256×256 and the input image channel is a single channel, it is necessary to preprocess face images in visible and infrared. Firstly, the images of the two datasets is cropped and the face region is reserved. The size of the cropped face image is 256×256. Then, the visible image is transformed into gray scale and normalized while the infrared image is enhanced and normalized.

The experimental hardware platform was a PC computer equipped with an Intel I9-9900K central processor and an NVIDIA RTX 2080 SUPER graphics card. During bidirectional cross-spectral face image conversion, 200 epochs were trained on the proposed network using the deep learning framework of PyTorch. The batch size was set to 1, and the learning rate of the first 100 epochs was 0.0002. Large search space was used to ensure network learning ability. In the last 100 epochs, the learning rate attenuated to 0 according to a linear law, and the weights were more finely adjusted to reduce the oscillation in training, so as to ensure a stable training of the network and obtain the optimal weights. Face feature extraction models are pre-trained on the WebFace face dataset, and then the CASIA and TINDERS multispectral face datasets are used to transfer-learn the network model. At the score fusion stage, the IDE of MATLAB was used to conduct adaptive weighted fusion experiments on the matching matrices.

## 3.2 Performance Metrics

In order to evaluate the cross-spectral face recognition performance of the proposed approach in this paper, we choose a comprehensive set of performance metrics which include: Genuine Acceptance Rate (GAR), False Acceptance Rate (FAR), Equal Error Rate (EER), d-prime and Area Under the Curve of ROC (AUC).

Since the bidirectional conversion network (i.e., BCSC-GAN) proposed in this paper is devised based on generative adversarial networks, the image quality of the bidirectionally generated faces of the network needs to be evaluated too, so the metrics of Fréchet Inception Distance (FID) is selected to evaluate the quality of generated face images [52]. Other common image quality metrics, such as PSNR and SSIM are effective for super-resolution, denoising and other tasks that requires image alignment first, but they are not good for quality evaluation between the input image and the generated image without registration. Therefore, FID has better discriminant ability and is an excellent quality index for GAN based methods. FID is calculated as:

$$FID(x,y) = \|\mu_x - \mu_y\|_2^2 + Tr\left[\Sigma_x + \Sigma_y - 2(\Sigma_x \cdot \Sigma_y)^{\frac{1}{2}}\right] \quad (10)$$

where $\|\bullet\|_2^2$ represents the square of L$_2$ norm and $Tr[\bullet]$ is the trace of the matrix; $x$ and $y$ denote the real image and generated image, respectively. $\mu_x$ and $\mu_y$ stand for the mean values of real image features and generated image features, respectively. $\Sigma_x$ and $\Sigma_y$ are the covariance matrices of real image features and generated image features, respectively.

**3.3 Face Conversion Experiments**

This section conducts comparative experiments on the aforementioned face conversion algorithms on different datasets in terms of the evaluation metrics introduced in Section 3.2.

Since the case of direct matching without image conversion (as many other methods in the literature do) and the CpAE network based on common potential subspace do not involve the image generation process, there is no need to compare FID for them. Instead, direct VIS-to-NIR face matching is used for cross-spectral face matching experiment. When other image conversion networks such as Pix2pix, CycleGAN and CSGAN are used, the quality of cross-spectral generated images are evaluated by FID at first. Since the three networks are designed for one-way image conversion, we train and use them twice, one for conversion of IR to Vis and the other for conversion of Vis to IR, reversely.

Table 4 shows the comparison results of face recognition performance and generated image quality between different image conversion methods and matching methods on the CASIA dataset. As can be seen from Table 4, for the CASIA dataset, CycleGAN has the best FID value in face conversion experiment of IR faces, and the proposed BCSC-GAN in this paper achieves the second best. In the subsequent matching experiment, the GARs for BCSC-GAN in the VIS band was 95% at $10^{-3}$ and 100% at $10^{-1}$, and EER was 2.42% and d-prime was 4.59. In the NIR band, the matching rate was 96% at $10^{-3}$ and 100% at $10^{-1}$, with EER value of 2.38% and d-prime value of 4.63. These results clearly demonstrate BCSC-GAN achieves the highest performance in terms of GAR, EER and d-prime under both bands of VIS and IR on the CASIA dataset, compared with other face conversion networks and matching methods.

Table 4. Comparison of recognition performance and generated image quality between different methods on the CASIA dataset.

| Method | Matching Manner | GAR(%) @FAR= $10^{-1}$ | GAR(%) @FAR= $10^{-3}$ | EER (%) | d-prime | FID |
|---|---|---|---|---|---|---|
| No Face Conversion | VIS vs NIR | 96.67 | 86.67 | 5.24 | 3.82 | - |
| CpAE[18] | VIS vs NIR | 88 | 76 | 11.82 | 3.36 | - |
| Pix2pix [26] | VIS vs VIS' | 94 | 66 | 6.18 | 3.11 | 51.22 |
|  | NIR vs NIR' | 94 | 58 | 8.27 | 2.99 | 59.05 |
| CycleGAN[29] | VIS vs VIS' | 100 | 92 | 4.40 | 4.22 | 48.59 |
|  | NIR vs NIR' | 100 | 92 | 2.42 | 4.33 | **36.61** |
| CSGAN[24] | VIS vs VIS' | 100 | 55 | 3.67 | 3.54 | 54.50 |
|  | NIR vs NIR' | 100 | 52 | 4.38 | 3.45 | 44.53 |
| **BCSC-GAN** (No Fusion) | VIS vs VIS' | **100** | **95** | **2.42** | **4.59** | **44.42** |
|  | NIR vs NIR' | **100** | **96** | **2.38** | **4.63** | 39.40 |

Table 5. Comparison of recognition performance and generated image quality between different methods on the TINDERS dataset at 1.5m standoff.

| Method | Matching Manner | GAR(%) @ FAR= $10^{-1}$ | GAR(%) @ FAR= $10^{-3}$ | EER (%) | d-prime | FID |
|---|---|---|---|---|---|---|
| No Image Conversion | VIS vs SWIR | 91.41 | 14.84 | 9.49 | 2.44 | - |
| CpAE[18] | VIS vs SWIR | 98.75 | **31.88** | 8.33 | **3.09** | - |
| Pix2pix [26] | VIS vs VIS' | 83.59 | 24.22 | 11.10 | 2.51 | 58.65 |
|  | SWIR vs SWIR' | 68.75 | 21.09 | 19.03 | 1.76 | 91.60 |
| CycleGAN[29] | VIS vs VIS' | 93.75 | 13.28 | **7.59** | 2.69 | 44.87 |
|  | SWIR vs SWIR' | 92.97 | 28.13 | 8.48 | 2.64 | 56.88 |
| CSGAN[24] | VIS VS VIS' | 62.50 | 25.34 | 17.41 | 1.96 | 43.15 |
|  | SWIR VS SWIR' | 70.31 | 9.38 | 17.13 | 1.93 | 56.84 |
| **BCSC-GAN** (No Fusion) | VIS vs VIS' | **99.22** | 25.78 | 8.31 | 2.73 | **40.17** |
|  | SWIR vs SWIR' | **95.31** | **41.41** | **5.47** | 2.90 | **52.30** |

Table 5 shows the comparison results of face recognition performance and generated image quality between different image conversion methods and matching methods on the TINDCERS dataset (1.5m). In the quality evaluation experiment of cross-spectral face conversion, the FID value of the proposed network of BCSC-GAN is the lowest, indicating the highest quality of face generation. In the matching experiment, the matching rate of BCSC-GAN under the VIS band is 25.78% at $10^{-3}$, 99.22% at $10^{-1}$ with an EER of 8.31% and d-prime of 2.73. In SWIR band, the matching rate was 41.41% at $10^{-3}$, 95.31% at $10^{-1}$ while EER was 5.47% and d-prime was 2.90. These results once again prove that our BCSC-GAN network yields the best results of bidirectional face conversion and cross-spectral face matching under both the VIS and SWIR bands on the TINDERS_SWIR 1.5m dataset.

Table 6. Comparison of recognition performance and generated image quality between different methods on the TINDERS dataset at 50m standoff.

| Method | Matching Manner | GAR(%) @FAR= $10^{-1}$ | GAR(%) @FAR= $10^{-3}$ | EER (%) | d-prime | FID |
|---|---|---|---|---|---|---|
| No Image Conversion | VIS vs SWIR | 39.5 | 2 | 33.25 | 0.98 | - |
| CpAE[18] | VIS vs SWIR | 61 | 0.5 | 25.54 | 1.61 | - |
| Pix2pix [26] | VIS vs VIS' | 55.47 | 0.78 | 24.33 | 1.38 | 85.89 |
| | SWIR vs SWIR' | 27.34 | 0.78 | 39.40 | 0.48 | 141.65 |
| CycleGAN[29] | VIS vs VIS' | 78 | 20 | 15.93 | 2.07 | 55.72 |
| | SWIR vs SWIR' | 35.5 | 2.5 | 34.39 | 0.79 | 226.15 |
| CSGAN[24] | VIS VS VIS' | 48 | 10.5 | 25.43 | 1.40 | 47.97 |
| | SWIR VS SWIR' | 37 | 1.5 | 37.18 | 0.86 | 93.59 |
| **BCSC-GAN** (No Fusion) | VIS vs VIS' | **87.5** | **48** | **12** | **2.36** | **44.10** |
| | SWIR vs SWIR' | **69** | **20** | **18.93** | **1.81** | **65.53** |

Table 6 compares the face recognition performance and generated image quality of different image conversion methods on the TINDERS_SWIR 50m dataset. In the evaluation experiment of cross-spectral image quality, compared with other image conversion methods, the FID value of the proposed network is the minimum, so the generation effect is the best. In the matching experiment, due to the large gap in the quality of the original cross-spectral images, the recognition rate of direct cross-spectral recognition is extremely low. Therefore, it is necessary to use the image transformation network to extract features from the image to the common potential subspace or the image transformation method. In the VIS band, the matching rate of BCSC-GAN is 48% at $10^{-3}$ and 87.5% at $10^{-1}$, with EER value of 12% and d-prime value of 2.36. In SWIR band, the matching rate was 20% at $10^{-3}$ and 69% at $10^{-1}$, EER value was 18.93% and d-prime value was 1.81. The results show that BCSC-GAN achieves the best GAR, EER and d-prime under both VIS and SWIR bands on the TINDERS_SWIR 50m dataset which clearly demonstrates its advantage over other methods.

Table 7. Comparison of recognition performance and generated image quality between different methods on the TINDERS dataset at 106m standoff.

| Method | Matching Manner | GAR(%) @FAR= $10^{-1}$ | GAR(%) @FAR= $10^{-3}$ | EER (%) | d-prime | FID |
|---|---|---|---|---|---|---|
| No Image Conversion | VIS vs SWIR | 26.5 | 3.5 | 42.32 | 0.54 | - |
| CpAE[18] | VIS vs SWIR | 62 | 11 | 26.43 | 1.38 | - |
| Pix2pix [26] | VIS vs VIS' | 61 | 6 | 23.07 | 1.46 | 78.78 |
| | SWIR vs SWIR' | 18.5 | 2 | 36.18 | 0.56 | 120.07 |
| CycleGAN [29] | VIS vs VIS' | 62.5 | 3 | 22.04 | 1.54 | 51.27 |
| | SWIR vs SWIR' | 36 | 2 | 36.14 | 0.90 | **87.39** |
| CSGAN[24] | VIS VS VIS' | 62.5 | 6 | 41.64 | 0.68 | **45.95** |
| | SWIR VS SWIR' | 25.5 | 1.5 | 36.46 | 0.68 | 99.02 |
| **BCSC-GAN** (No Fusion) | VIS vs VIS' | **70.5** | **28.5** | **18.11** | **1.67** | 53.25 |
| | SWIR vs SWIR' | **40** | **5** | **27.39** | **1.25** | 94.94 |

Table 7 compares the face recognition performance and generated image quality of different image conversion methods on TINDERS_SWIR 106m dataset. In the evaluation experiment of the quality of cross-spectral generated images, compared with CSGAN and CycleGAN networks, the FID value of the proposed network is sub-small, and the comprehensive generation effect is the best in the two bands. In the matching experiment, the matching rate of the proposed network of BCSC-GAN in this paper is 28.5% at $10^{-3}$ and 70.5% at $10^{-1}$ in VIS band, EER value is 18.11% and d-prime value is 1.67. In SWIR band, the matching rate was 5% at $10^{-3}$ and 40% at $10^{-1}$, with EER value of 27.39% and d-prime value of 1.25. The results show that BCSC-GAN achieves the best GAR, EER and d-prime under both VIS and SWIR bands on the TINDERS_SWIR 106m dataset which once again proves its superiority over other methods.

### 3.4 Score Fusion Experiment

After completing the bidirectional conversion experiments and the matching experiments, we this subsection continue to carry out score fusion experiments using our proposed SAWF strategy in Section 4.3.1. The fusion experiments are carried out based on the following four groups of experimental data and the results are presented respectively.

Table 8 shows the comparison results between different fusion rules in terms of GAR, EER, d-prime and AUC on CASIA dataset. As can be seen from the experimental results in Table 8, the performance metrics of GAR, EER and d-prime on the CASIA dataset after matching score fusion using the SAWF rule were significantly better than those without fusion (See Table 4). At the same time, the AUC metrics is used to evaluate each fusion rule. Experimental results show that the performance evaluation metrics of the proposed SAWF rule are all higher than those of other commonly-used fusion rules, including the maximum value fusion rule, the geometric average fusion rule and the arithmetic fusion rule.

Tables 9, 10 and 11 show the comparison results of matching scores for different fusion rules at different acquisition distances on the TINDERS dataset, from 1.5m to 50m and 106m.

Table 8. The comparison results between different fusion rules in terms of GAR, EER, d-prime and AUC on the CASIA dataset.

| Fusion Rule | GAR(%) at FAR= $10^{-1}$ | GAR(%) at FAR= $10^{-3}$ | EER(%) | d-prime | AUC |
|---|---|---|---|---|---|
| Maximum value | 100 | 95 | 2.06 | 5.02 | 0.99945 |
| Geometric average | 100 | 97 | 1.96 | 4.38 | 0.99960 |
| Arithmetic average | 100 | 97 | 1.92 | 5.16 | 0.99962 |
| **SAWF** | **100** | **100** | **0.0001** | **6.54** | **1.00000** |

Table 9. The comparison results between different fusion rules in terms of GAR, EER, d-prime and AUC on the TINDERS 1.5m dataset.

| Fusion Rule | GAR (%) at FAR= $10^{-1}$ | GAR (%) at FAR= $10^{-3}$ | EER(%) | d-prime | AUC |
| --- | --- | --- | --- | --- | --- |
| Maximum value | 99.22 | 27.34 | 8.26 | 2.75 | 0.97538 |
| Geometric average | 100 | 75.78 | 2.85 | 3.21 | 0.99731 |
| Arithmetic average | 100 | 83.59 | 2.18 | 3.63 | 0.99892 |
| **SAWF** | **100** | **85.16** | **1.62** | **4.15** | **0.99939** |

Table 10. The comparison results between different fusion rules in terms of GAR, EER, d-prime and AUC on the TINDERS 50m dataset.

| Fusion Rule | GAR (%) at FAR= $10^{-1}$ | GAR (%) at FAR= $10^{-3}$ | EER(%) | d-prime | AUC |
| --- | --- | --- | --- | --- | --- |
| Maximum value | 87.5 | 46.5 | 11.93 | 2.49 | 0.95891 |
| Geometric average | 92 | 39 | 9.5 | 2.58 | 0.97032 |
| Arithmetic average | 91 | 47.5 | 9.5 | 2.59 | 0.97711 |
| **SAWF** | **98** | **58.5** | **7.07** | **3.40** | **0.98741** |

Table 11. The comparison results between different fusion rules in terms of GAR, EER, d-prime and AUC on the TINDERS 106m dataset.

| Fusion Rule | GAR (%) at FAR= $10^{-1}$ | GAR (%) at FAR= $10^{-3}$ | EER(%) | d-prime | AUC |
| --- | --- | --- | --- | --- | --- |
| Maximum value | 70.5 | 28.5 | 18 | 1.7814 | 0.90540 |
| Geometric average | 77.5 | 25 | 15.57 | 1.956 | 0.93088 |
| Arithmetic average | 86 | 43 | 12.39 | 2.2423 | 0.94885 |
| **SAWF** | **88.5** | **42.5** | **11.07** | **2.5462** | **0.95541** |

As can be seen from the experimental results of Table 9, Table 10 and Table 11, for the TIEDRS dataset, the GAR, EER and d-prime after the matching score fusion using the SAWF rule significantly exceeded the results before the fusion (see Table 5, Table 6 and Table 7). Once again, experimental results show that the performance evaluation metrics of the proposed fusion rule are all better than those of other commonly-used fusion rules, i.e., the maximum value fusion rule, the geometric average fusion rule and the arithmetic fusion rule. These results once more prove the superiority of our proposed score fusion rule of SAWF.

The SAWF algorithm based on GAR and d-prime takes into account the GAR and d-prime values of the matching scores in the single spectrum after conversion, so that the matching score after fusion not only averages the error, but also ensures the distance between the intra-class features and inter-class. Therefore, the final decision result is a globally optimized decision.

## 4. Conclusion

This paper studied the problem of cross-spectral face recognition and proposed a new bidirectional face conversion approach which reduces the more complex problem of cross-spectral recognition to the simpler intra-spectral recognition problem. Besides, the bidirectional face conversion approach enables fusion of bidirectional matching scores which helps further improving the cross-spectral recognition rate. The proposed network (BCSC-GAN) features a cyclical synthesis GAN structure and an Identity Retaining module which together not only guarantee high quality of generated face images but also high recognition performance. A new composite loss function is designed to overcome the modal differences caused by different spectral characteristics. Additionally, we proposed a Self-Adaptive Weighted Fusion (SAWF) strategy for score-level fusion to further boost the fusion effect. In terms of performance metrics such as FID, recognition rate, equal error rate and normalized distance, experimental results on different datasets show that our proposed network is superior than other state-of-the-art methods. It is also observed that our approach is better than the unfused case and other commonly-used traditional fusion rules, which further justifies the effectiveness and superiority of the proposed bidirectional conversion approach.